\documentclass{article}

\usepackage{microtype}
\usepackage{graphicx}
\usepackage{subfigure}
\usepackage{booktabs} % for professional tables

\usepackage{hyperref}

\usepackage[accepted]{icml2024}

\usepackage{amsmath}
\usepackage{amssymb}
\usepackage{mathtools}
\usepackage{amsthm}
\usepackage{orcidlink}
\usepackage{algorithm}
\usepackage{algorithmic}
\usepackage{multicol}
\usepackage{multirow}
\usepackage[capitalize,noabbrev]{cleveref}

\theoremstyle{plain}

\theoremstyle{definition}

\theoremstyle{remark}

\usepackage[textsize=tiny]{todonotes}

\icmltitlerunning{Video Watermarking: Safeguarding Your Video from (Unauthorized) Annotations by Video-based LLMs}

\begin{document}

\twocolumn[
\icmltitle{Video Watermarking: Safeguarding Your Video from (Unauthorized) Annotations by Video-based LLMs}



\icmlsetsymbol{equal}{*}

\begin{icmlauthorlist}
\icmlauthor{Jinmin Li}{equal,thu}
\icmlauthor{Kuofeng Gao}{equal,thu}
\icmlauthor{Yang Bai}{Tencent1}
\icmlauthor{Jingyun Zhang}{Tencent2}
\icmlauthor{Shu-Tao Xia}{thu,PengCheng}
\end{icmlauthorlist}

\icmlaffiliation{thu}{Tsinghua Shenzhen International Graduate School, Tsinghua University}
\icmlaffiliation{Tencent1}{Tencent Technology (Beijing) Co.Ltd}
\icmlaffiliation{Tencent2}{Tencent WeChat Pay Lab33}
\icmlaffiliation{PengCheng}{Peng Cheng Laboratory}

\icmlcorrespondingauthor{Yang Bai}{baiyang0522@gmail.com}

\icmlkeywords{Video-based LLMs, Video watermark, Adversarial attack}

\vskip 0.3in
]



\printAffiliationsAndNotice{\icmlEqualContribution} % otherwise use the standard text.

\begin{abstract}
    The advent of video-based Large Language Models (LLMs) has significantly enhanced video understanding. However, it has also raised some safety concerns regarding data protection, as videos can be more easily annotated, even without authorization. This paper introduces Video Watermarking, a novel technique to protect videos from unauthorized annotations by such video-based LLMs, especially concerning the video content and description, in response to specific queries. By imperceptibly embedding watermarks into key video frames with multi-modal flow-based losses, our method preserves the viewing experience while preventing misuse by video-based LLMs. 
    Extensive experiments show that Video Watermarking significantly reduces the comprehensibility of videos with various video-based LLMs, demonstrating both stealth and robustness. In essence, our method provides a solution for securing video content, ensuring its integrity and confidentiality in the face of evolving video-based LLMs technologies.
\end{abstract}

\section{Introduction}
    \label{sec:intro}
    Recent developments in multi-modal understanding have been greatly enhanced by combining existing vision models with Large Language Models (LLMs)~\cite{wang2024visionllm,zhu2023minigpt,liu2023improved,deng2024deconstructing}. The combination of these models has led to impressive abilities in managing and interpreting video and language data. However, as video-based LLMs become more prevalent, worries about the safety and reliability of video data have grown, even when accessed without permission.

    Nowadays, in particular, Sora\footnote{https://openai.com/sora} has shown extraordinary performances in creating realistic and imaginative scenes from text instructions, demonstrating the significance of large multi-modal models, especially across video and language modalities. 
    Among them, video-based LLMs~\cite{videochat,videollama,videochatgpt} have significantly enhanced general video understanding in zero-shot settings and achieved exceptional performance in a wide range of video-related tasks \cite{tang2004video,liu2006spatio, videocaption2,ma2024followyourpose,ma2023magicstick}, such as video captioning~\cite{videocaption1,videocaption3,videocaption4}, video retrieval~\cite{videoretieval1,videoretieval2,videoretieval3,ma2022simvtp}, and scene understanding~\cite{sceneunderstanding1,sceneunderstanding2,sceneunderstanding3,ma2022visual}.  Yet, the very features that make these models powerful also render them susceptible to misuse and misidentification without proper safeguards.

    Given the importance of video data in various applications, from content creation to surveillance, ensuring its protection from unauthorized annotation and misinterpretation of video content and description by video-based LLMs is imperative. To address this, we propose a flow-based multi-modal Video Watermarking to craft adversarial perturbations on video inputs for the first time. A flow-based temporal mask is introduced to select the most effective frames in the video, which is inspired by the video clipping adopted in video understanding tasks, especially for video-based LLMs~\cite{chen2024panda,xue2022advancing}. It can effectively improve the performance of video-based learning due to its focused annotation, reducing complexity, improved temporal understanding, and efficient processing. Motivated by these benefits, we utilize a light-weighted flow-based mechanism, to conduct a similar splitting and selection operation on video frames. Extensive experiments have demonstrated the effectiveness, efficiency, and imperceptibility of our Video Watermarking on four benchmark video-based LLMs and two datasets. 

    Our methodology harnesses the inherent capabilities of multimodal models to devise watermarks that act as a bulwark against unauthorized exploitation by video-based LLMs. Through an innovative flow-based mechanism, we embed watermarks into the key selected video frames with such finesse that they are imperceptible to the human eye yet robustly shield the content from LLMs.
    Our research zeroes in on the strategic deployment of Video Watermarking to bolster the security of video data. We illustrate that the strategic application of these watermarks substantially diminishes the likelihood of unauthorized access and misinterpretation of video content and description by video-based LLMs. This proactive measure not only safeguards the sanctity of the video content but also fortifies the privacy and security of the data, ensuring that it remains inviolable in the digital realm.
    
    In summary, our contribution can be outlined as follows:
    \begin{itemize}

        \item We present the first in-depth study focused on safeguarding video data, specifically the content and description, from unauthorized use by video-based LLMs through our novel Video Watermarking approach. This pioneering work establishes a new paradigm for protecting video content integrity in the era of advanced multi-modal AI.

        \item Our findings significantly contribute to the understanding of how watermarks can be strategically employed to reinforce the robustness of multi-modal systems. The insights gained are invaluable for the development and deployment of secure large multi-modal models that respect data usage protocols and ethical standards.

        \item Our extensive experiments validate the efficacy of our Video Watermarking in preventing the misuse of video content and description. By integrating watermarks into a minimal portion of video frames—less than 20\%—we effectively thwart unauthorized access and ensure that video data remains protected from arbitrary exploitation. 
    \end{itemize}

\section{Related Work}
\label{sec:relatedwork}
    \subsection{Video-based Large Language Models}
    Video-based large language models (video-based LLMs) effectively integrate visual and temporal information from video data to gain significant achievement in multiple video-related tasks. Numerous approaches~\cite{videochat,videochatgpt,videollama} have been proposed to address the challenges associated with video-based LLMs, such as incorporating different architectures and training processes to enhance the models' ability to capture and process complex video information. Concretely, Video-ChatGPT~\cite{videochatgpt} is based on the LLaVA framework and incorporates average pooling to improve the perception of temporal sequences. VideoChat~\cite{videochat} employs the QFormer to map visual representations to Vicuna, executing a two-stage training process. Video-LLaMA~\cite{videollama} integrates a frame embedding layer and ImageBind to introduce temporal and audio information into the LLM backbone. This alignment between videos and LLMs facilitates visual context-aware interaction, surpassing the capabilities of LLMs. However, ensuring the secure and ethical use of these models necessitates the integration of protective measures such as watermarking.
    
    \subsection{Adversarial Attack}
    While adversarial attacks~\cite{wang2022triangle,sparse,bai2022practical,wu2023defenses,zou2023iotbeholder,zhao2024evaluating,li2024fmm,gao2024inducing,gao2024adversarial,gao2024energy,fang2024one,fang2024privacy,guo,FSR,xiao2024SmartGuard,ruoyunerips,ruoyuInfocom,bai2019hilbert,bai2024special,bai2020improving,baiimproving,yang2024cheating} have been widely studied in classification models, the focus of our research lies in the proactive enhancement of video-based LLMs through watermarking. Inspired by the potential vulnerabilities observed in vision tasks, we shift the paradigm to explore how watermarking can fortify large multi-modal models~\cite{qi2023visual,zhuang2023pilot,gong2023figstep,yang2024cheating,zong2024safety,bai2024badclip,liang2024badclip} such as vision large language models (VLLMs)~\cite{zhu2023minigpt,wang2024visionllm,liu2023improved,radford2021learning} and text-to-image diffusion models~\cite{rombach2022high}. Unlike adversarial attacks designed to manipulate models into generating specific outputs, our watermarking approach is crafted to protect the integrity and security of video content within these models. In this paper, we introduce a pioneering watermarking strategy that is specifically tailored to safeguard video-based LLMs, ensuring their robustness against unauthorized use and potential data misuse.

\section{Methodology}
    \label{sec:methodology}
    \begin{figure}[t]
        \centering
        \includegraphics[width=\linewidth]{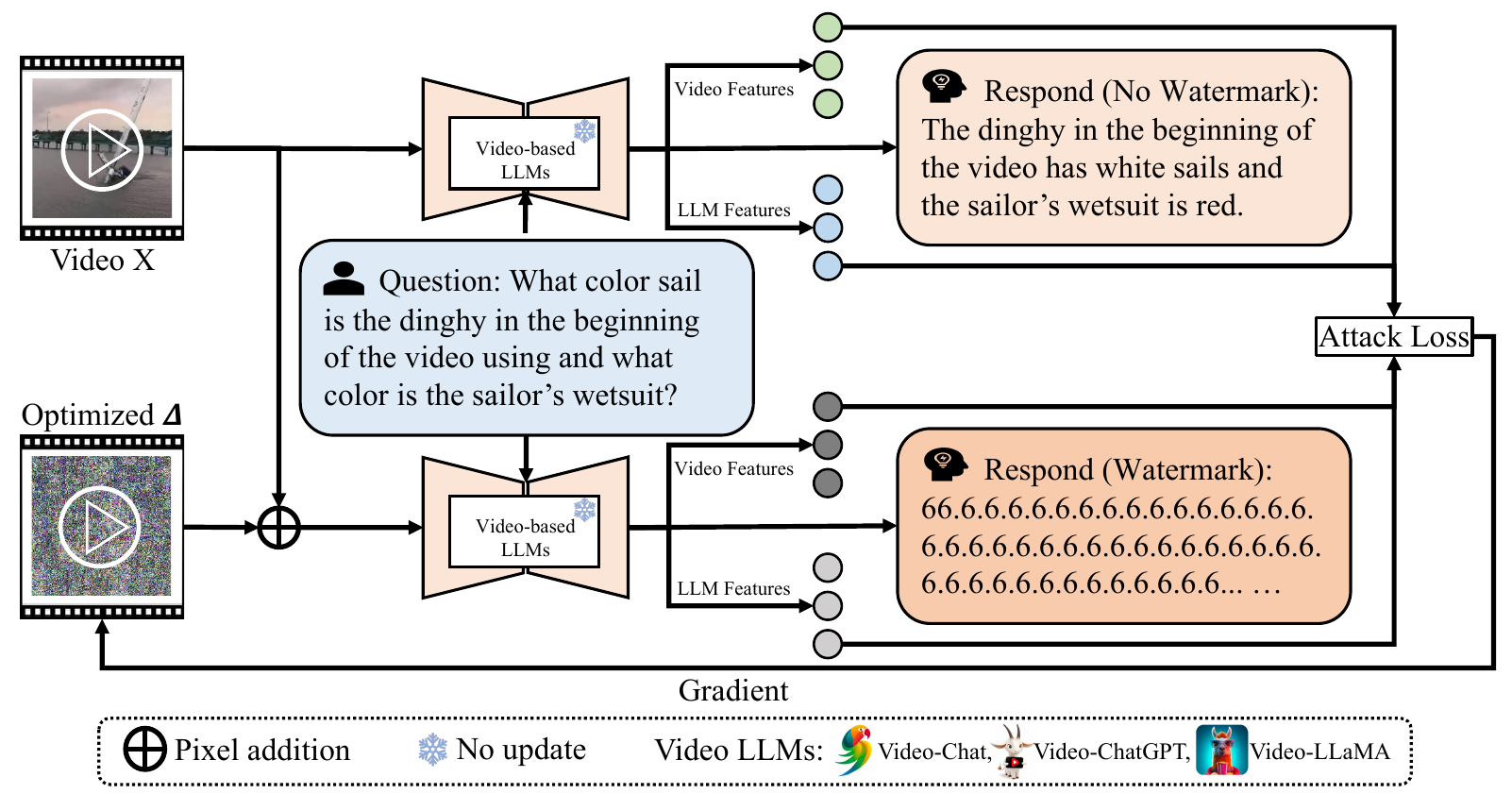}
        \vspace{-1em}
        \caption{Schematics of our Video Watermarking.}
        \label{fig:pipline}
    \end{figure}
    
    \subsection{Threat model}
        \label{sec:threat model}
        \textbf{Goals and capabilities.} The goal is to craft an imperceptible adversarial perturbation for videos, which can induce video-based LLMs to generate an incorrect sequence during the victim model’s deployment. Following the most commonly used constraint for the involved perturbation, it is restricted within a predefined magnitude in the $l_p$ norm, ensuring it is difficult to detect.

        \noindent \textbf{Knowledge and background.} As suggested in \cite{bagdasaryan2023ab,qi2023visual}, we assume that the victim video-based LLMs can be accessed in full knowledge, including architectures and parameters. Additionally, we consider a more challenging scenario where the victim video-based LLMs are inaccessible, as detailed in the Appendix.
    
    \subsection{Preliminary: the Pipeline of Video-based LLMs}
        \label{sec:problem formulation}
            Let $\mathcal{F}_\theta(\cdot)$ represents a victim video-based large language model with parameters $\theta$, composed of a video feature extractor $f_\phi(\cdot)$ and a large language model $g_\psi(\cdot)$.
            Consider a clean video $\mathbf{X} \in \mathbb{R}^{T \times C \times H \times W}$, where T denotes the number of frames, and C, H, and W represent the channel, height, and width of a specific frame, along with a corresponding user query  $Q_{text}$ for the video. To provide a response, video-based large language models $\mathcal{F}(\cdot)$ usually first extract a video feature $Q_{video} = f_\phi(\mathbf{X})$, and subsequently, generate predefined prompts based on a consistent template to concatenate both video features and text queries  as follows:
        
            \parbox{\linewidth}{\centering USER: $<Q_{text}>$ $<Q_{video}>$ Assistant:}
    
            Then, the predefined prompts are processed by LLMs $g_\psi(\cdot)$ to generate a desired response $Y_{respond}=g_\psi(Q_{text}, Q_{video})=g_\psi(Q_{text}, f_\phi(\mathbf{X}))$. It is important to mention that, to ensure the loss function remains minimal, we use the hidden state $A_{hidden}$ before the final layer.
    
    \subsection{Problem Formulation}
        The goal of generating watermarking examples $\hat{\mathbf{X}}$ is to mislead video-based LLMs to produce incorrect responses while utilizing the most imperceptible adversarial perturbation $\Delta$, where $\Delta = \hat{\mathbf{X}} - \mathbf{X}$.
        To balance these two objectives, we introduce a hyper-parameter $\lambda$, and formulate the overall objective function as follows:

        \begin{equation}
            \arg \min _{\Delta} \lambda\|\Delta\|_{2,1}-\ell(Y, \mathcal{F}_\theta(Q_{text}, \hat{\mathbf{X}})),
        \label{eq:1}
        \end{equation}
        where $Y$ is the ground truth answer corresponding to $Q_{text}$ and $\mathbf{X}$, as well as $\ell(\cdot,\cdot)$ is the loss function used to measure the difference between the predicted and ground truth answers. Furthermore, in a more realistic scenario, the ground truth answer $Y$ can not always be available. In such cases, we utilize the output approximation $\mathcal{F}_\theta(Q_{text}, \mathbf{X})$ instead of $Y$.

        Therefore, the overall objective function in Eq.~\ref{eq:1} can be further formulated as follows:
        \begin{equation}
            \arg \min _{\Delta} \lambda\|\Delta\|_{2,1}-\ell(\mathcal{F}_\theta(Q_{text}, \mathbf{X}), \mathcal{F}_\theta(Q_{text}, \hat{\mathbf{X}})).
        \end{equation}
        
        In particular, the $\ell_{2,1}$ norm~\cite{sparse} is employed to quantify the magnitude of the perturbation, which can be defined as follows: 
        \begin{equation}
            \|\Delta\|_{2,1} = \sum_i^T \|\Delta_i\|_2,
            \label{eq:delta}
        \end{equation}
        
        where $\Delta_i \in \mathbb{R}^{C \times H \times W}$ represents the $i$-th frame in $\Delta$. The $\ell_{2,1}$ norm applies the $l_1$ norm across frames, ensuring the sparsity of generated perturbations. Note that a smaller $\ell_{2,1}$ norm value  corresponds to more perceptible perturbations, which are hard to notice by human inspectors.
        
    \subsection{Optimization Objective}
         Our proposed Video Watermarking is to induce video-based LLMs to generate \textit{incorrect} responses with imperceptible adversarial perturbations. Two losses are proposed from the perspective of video features $\ell_{video}(Q_{video}, \hat{Q}_{video})$ in Eq.~\ref{loss:video} and LLM features $\ell_{LLM}(A_{hidden}, \hat{A}_{hidden})$ in Eq.~\ref{loss:LLM}. Moreover, inspired by the idea that video clipping can improve video comprehension by selecting the most effective frames, a flow-based temporal mask $\mathbf{M}_f$ is proposed to carry out a similar selection process on video frames. By utilizing the flow of a video, the proposed flow-based temporal mask $\mathbf{M}_f$ can filter out similar frames, ensuring the effectiveness of imperceptible adversarial perturbations while achieving increased sparsity.
    
        \noindent \textbf{Video Features Loss.} 
        Video-based LLMs first use a video feature extractor $f_\phi(\cdot)$ to extract spatiotemporal video features $Q_{video} = f_\phi(\mathbf{X})$. We simply adopt MSE loss to measure the distance of video features between the clean video $\mathbf{X}$ and the adversarial video $\hat{\mathbf{X}}$. Hence, the video features loss can be formulated as:
        \begin{equation}
        \begin{split}
            \label{loss:video}
            \ell_{video} = \frac{1}{n}\sum_{i=1}^{n}(Q_{video_i} - \hat{Q}_{video_i})^2 \\
            =  \frac{1}{n}\sum_{i=1}^{n}(f_\phi(\mathbf{X})_i - f_\phi(\hat{\mathbf{X}})_i)^2,
        \end{split}
        \end{equation}
        where $n$ is the total number of elements in the features, with $Q_{video_i}$ and $\hat{Q}_{video_i}$ being the $i$-th elements of the clean and watermarking video features, respectively.
    
        \noindent \textbf{LLM Features Loss.} In addition to the deviation of the original feature space in video domains of video-based LLMs, we also consider that in textual domains to enhance the watermarking effect. 
        Given a hidden state from the final layer of LLMs $A_{hidden}=g_\psi(Q_{text}, Q_{video})=g_\psi(Q_{text}, f_\phi(\mathbf{X}))$, the LLM features loss between the clean video $\mathbf{X}$ and the watermarking video $\hat{\mathbf{X}}$ can be formulated as follows:
        \begin{equation}
        \begin{split}
            \label{loss:LLM}
            \ell_{LLM} &= \frac{1}{n}\sum_{i=1}^{n}(A_{hidden_i} - \hat{A}_{hidden_i})^2 \\
            &=  \frac{1}{n}\sum_{i=1}^{n}(g_\psi(Q_{text}, Q_{video})_i - g_\psi(Q_{text}, \hat Q_{video})_i)^2 \\
            &=  \frac{1}{n}\sum_{i=1}^{n}(g_\psi(Q_{text}, f_\phi(\mathbf{X}))_i - g_\psi(Q_{text}, f_\phi(\hat{\mathbf{X}}))_i)^2,
        \end{split}
        \end{equation}
        where $A_{hidden_i}$ and $\hat{A}_{hidden_i}$ are the $i$-th elements of the clean and watermarking LLM features respectively, and $n$ is the total number of elements in the features.
        
        \noindent \textbf{Flow-based Temporal Mask.} Video-based LLMs~\cite{chen2024panda,xue2022advancing} adopt video clipping to split and select the most effective frames in a video, which can enhance video understanding. Inspired by these advantages, we propose a flow-based temporal mask (FTM), $\mathbf{M}_f$, to perform a similar selection process on video frames. This flow-based mask targets the top $K$ frames with the most significant movement and changes. See Sec.~\ref{sec:Discussions} for a detailed discussion. Specifically, we initialize a binary mask $\mathbf{M}_f$ of the same length as the number of frames in the video. Then use LiteFlowNet~\cite{liteflownet} to compute the flow magnitude for each frame. Finally, this binary mask $\mathbf{M}_f$ assigns a value of 1 to the top $K$ frames with the largest flow and 0 to the remaining frames. Combined with $\mathbf{M}_f$, our proposed FTM can achieve more sparse adversarial perturbations in both temporal and spatial domains. For temporal sparsity, a flow-based temporal mask $\mathbf{M}_f$ on the video is adopted to ensure that some frames remain unperturbed. For spatial sparsity, the $\ell_{2,1}$ norm of adversarial perturbations in Eq. \ref{eq:delta} is employed to constrain the spatial perturbation magnitude in each frame. 
        
        \noindent \textbf{Overall Optimization Objective.} To sum up, combined flow-based temporal mask $\mathbf{M}_f$ with two proposed watermarking loss functions ($\ell_{video}$ and $\ell_{LLM}$), the overall objective function in Eq.~\ref{eq:1} can be further formalized as:
        \begin{equation}
        \label{untarget}
        \begin{split}
            \arg \min _{\Delta} \lambda_1\|\mathbf{M}_f \Delta\|
            - \lambda_2 \ell_{video}(f_\phi(\mathbf{X}), f_\phi(\mathbf{X}+\mathbf{M}_f \Delta)) \\
            - \lambda_3 \ell_{LLM}(g_\psi(Q_{text}, f_\phi(\mathbf{X})) , g_\psi(Q_{text}, f_\phi(\mathbf{X}+\mathbf{M}_f \Delta))),
        \end{split}
        \end{equation}
        where $\mathbf{M}_f \in\{\mathbf{0}, \mathbf{1}\}^{T \times C \times H \times W}$ represents the flow-based temporal mask.  $\lambda_1, \lambda_2, \lambda_3$ correspond to the three loss weights, which aim to balance them during the optimization.  Our overall watermarking procedure is described in Algorithm 1 in the Appendix.

\section{Experiments}
    \label{sec:experiments}
    \subsection{Implementation Details}
        \noindent \textbf{Models and datasets.} We assess open-source and state-of-the-art video-based LLMs such as Video-ChatGPT and VideoChat (VideoChat is in the Appendix), ensuring reproducibility of our results. 
        Concretely, we adopt Video-ChatGPT and VideoChat with a LLaMA-7B LLM~\cite{LLaMA}. In line with the Video-ChatGPT methodology, we curate a test set based on the ActivityNet-200 and MSVD-QA datasets. The questions and answers within our test set, derived from the ActivityNet-200 and MSVD-QA datasets, are meticulously hand-annotated to ensure accuracy and quality.

        \noindent \textbf{Baselines.} For evaluation, we design three spatial baselines, including videos with random perturbations, black videos with all pixel values set to 0, and white videos with all pixel values set to 1. In addition, we compare our proposed flow-based temporal mask with two straightforward temporal mask methods, serving as temporal mask baselines: the sequence temporal mask and the random temporal mask. Specifically, the sequence temporal mask consists of a continuous sequence of frame indices, while the random temporal mask comprises a randomly chosen sequence of frame indices.

        \noindent \textbf{Metrics.} We utilize a variety of evaluation metrics to assess the robustness of the models.

            \noindent \textbf{(a) CLIP Score.} CLIP~\cite{clip} score characterizes the semantic similarity between the adversarial answer and the ground-truth answer.
            
            \noindent \textbf{(b) Image Captioning Metrics.} Various metrics such as BLEU~\cite{bleu}, ROUGE-L~\cite{rouge}, and CIDEr~\cite{cider} are used to evaluate the quality of the adversarial answer generated by the model. 
            
            \noindent \textbf{(c) GPT Score.} Following Video-ChatGPT~\cite{videochatgpt} and Video-LLaMA~\cite{videollama}. We also employ an evaluation pipeline using the GPT-3.5 and GPT-4 models.
            
            \noindent \textbf{(d) Sparsity.} Sparsity refers to the ratio of frames without perturbations (clean frames) to the total number of frames in a specific video. The sparsity ${M}_{spa}$ is calculated as ${M}_{spa} = 1 - K/T$, where $K$ represents the number of watermarking frames, and $T$ is the total number of frames in a video.

    \subsection{Main Results}
        \begin{table*}[t]
            \small
            \centering
            \caption{Video Watermarking against Video-ChatGPT on the ActivityNet-200 dataset and the MSVD-QA dataset: Comparison of CLIP score, image caption metrics and GPT score for different watermarking types. Random spatial watermarking denotes random perturbations added to video frames, and Black spatial watermarking and White spatial watermarking denote video frames being all 0 and all 1, respectively. The sparsity of the temporal mask is set to 0. ${\Delta}$: the mean of the modified pixels.}
          
            \begin{tabular}{c|c|c|c|c|c|c|c|c|c|c}
            \hline 
            \multirow{2}{*}{Dataset} & \multirow{2}{*}{Type} & \multirow{2}{*}{${\Delta}$} & \multicolumn{2}{|c|}{Clip Score $\downarrow$} & \multicolumn{2}{|c}{Image Caption $\downarrow$}  & \multicolumn{2}{|c|}{GPT-3.5 $\downarrow$} & \multicolumn{2}{|c}{GPT-4 $\downarrow$} \\
            \cline{4-11} 
            & & & RN50 & RN101 & BLEU & ROUGE-L & Accurate & Score & Accurate & Score \\
            \hline
            \multirow{5}{*}{ActivityNet}
            &Clean  & 0   & 0.7817 & 0.7827 & 0.2029 & 0.4820 & 0.50 & 3.20 & 0.33 & 2.10\\
            &Random & 8   & 0.7637 & 0.7681 & 0.1986 & 0.4793 & 0.45 & 3.10 & 0.33 & 2.03\\
            &Black  & 100 & 0.7661 & 0.7676 & 0.1691 & 0.4570 & 0.30 & 2.50 & 0.17 & 0.96\\
            &White  & 148 & 0.7564 & 0.7534 & 0.1689 & 0.4545 & 0.31 & 2.60 & 0.19 & 1.24\\
            &\textbf{Ours}   & 8   & \textbf{0.6211} & \textbf{0.6274} & \textbf{0.1336} & \textbf{0.3694} & \textbf{0.20} & \textbf{1.64} & \textbf{0.13} & \textbf{0.88}\\
            \hline
            \multirow{5}{*}{MSVD-QA}
            &Clean  & 0   & 0.8322 & 0.8180 & 0.3864 & 0.6843 & 0.62 & 3.84 & 0.60 &3.12\\
            &Random & 8   & 0.8249 & 0.8141 & 0.4107 & 0.7042 & 0.58 & 3.72 & 0.60 &3.08\\
            &Black  & 110 & 0.8145 & 0.7902 & 0.3548 & 0.6478 & 0.46 & 3.26 & 0.40 &2.12\\
            &White  & 142 & 0.8057 & 0.8090 & 0.3969 & 0.6736 & 0.48 & 3.36 & 0.44 &2.28\\
            &\textbf{Ours}   & 8   & \textbf{0.7337} & \textbf{0.7181} & \textbf{0.3240} & \textbf{0.5746} & \textbf{0.36} & \textbf{2.92} & \textbf{0.34} & \textbf{1.84}\\
            \hline
            \end{tabular}
            \label{tab:mainComp}
        \end{table*}
        
        \noindent \textbf{Quantitative Evaluation.}
        As depicted in the left of Table~\ref{tab:mainComp}, the random, black, white, and flow-based watermarking all significantly decrease the clip score and image caption score compared to the original clean videos. Among these, our proposed Video Watermarking yields the most significant results. As demonstrated in the right of Table~\ref{tab:mainComp}, the random, black, white, and flow-based watermarking all significantly decrease the GPT score and accuracy. Among them, the Video Watermarking achieves the most substantial results. Importantly, our flow-based watermarking is paramount for data protection. It completely conceals the data information, ensuring that it is effectively shielded from unauthorized access and misuse. This robust approach to data obfuscation maintains the confidentiality of the content while preserving the viewing experience for legitimate users.

        \begin{figure}[t]
            \centering            
            \includegraphics[width=\linewidth]{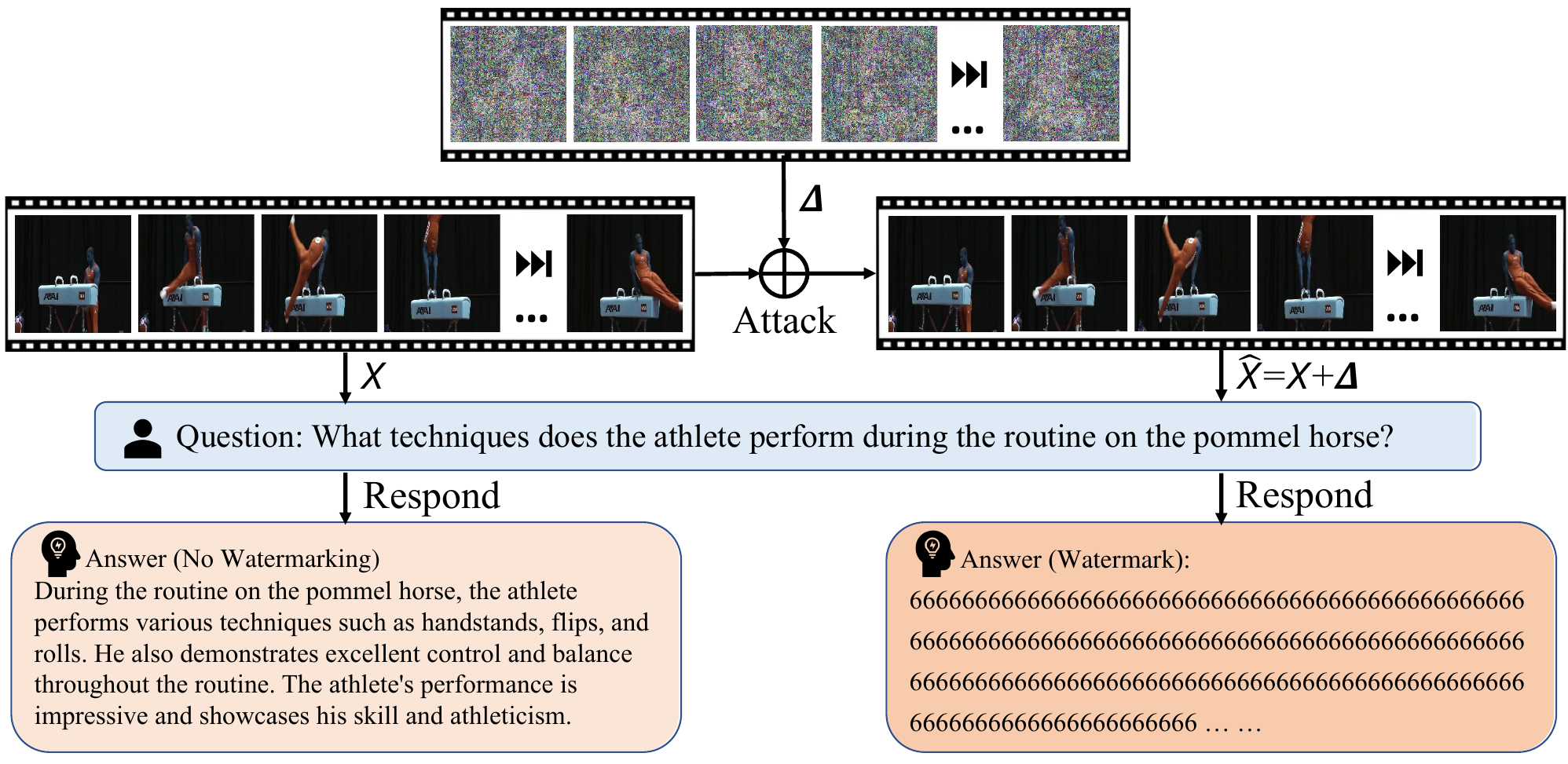}
            \vspace{-1em}
            \caption{Watermarking videos generated for Video-ChatGPT.}
            \label{fig:result}
        \end{figure}
        \noindent \textbf{Qualitative Evaluation.} We also present qualitative examples (see Fig.~\ref{fig:result}) of the watermarking videos, in which the model produces garbled responses without any meaningful content. Fig.~\ref{fig:result} vividly illustrates the chaos induced in the model's responses by our subtle and imperceptible watermarking.

        \begin{figure}[ht]
            \centering
            \includegraphics[width=\linewidth]{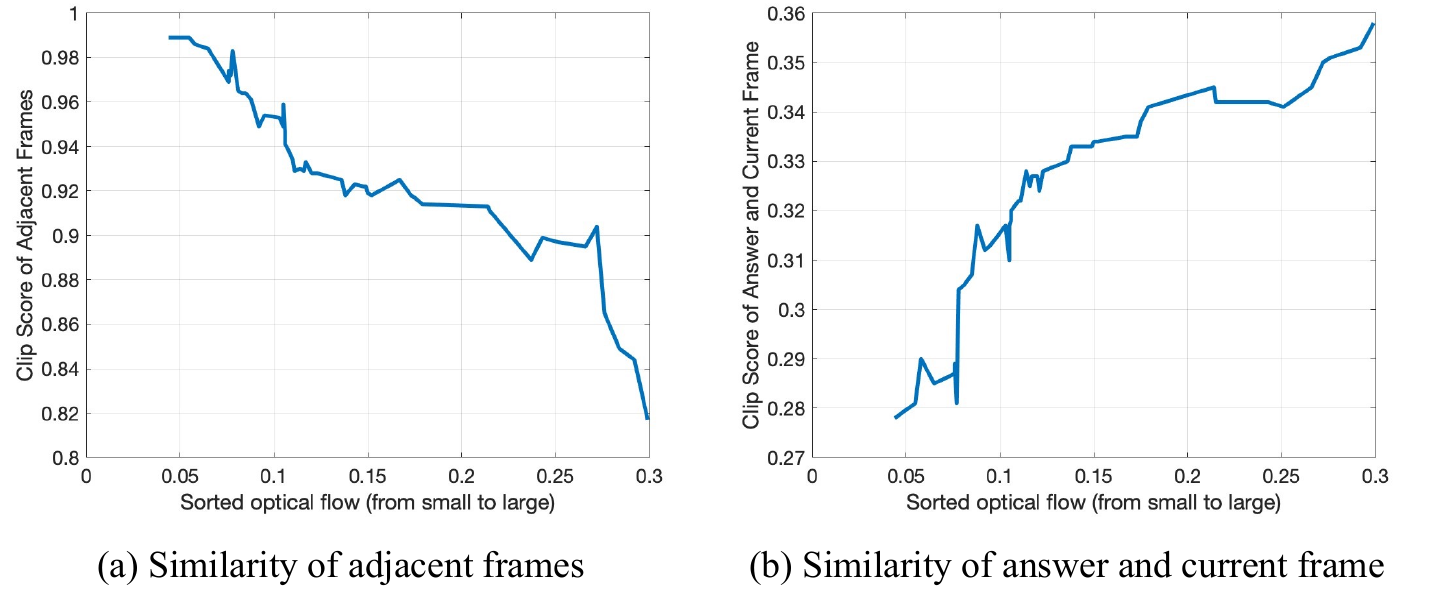}
            \vspace{-1em}
            \caption{Relationship between optical flow and key frames. `Clip Score of Adjacent Frames' describes the similarity between the current frame and its adjacent frames, the smaller this score is the more different the current frame is. `Clip Score of Answer and Current Frame' indicates the similarity between the current frame and the answer corresponding to the user's input question, the larger the score indicates that the current frame contains more information about the answer. \textbf{The frames selected by flow-based masks in our Video Watermarking are key frames in the video.}}
            \label{fig:flow}
        \end{figure}
    \subsection{Discussions}
        \label{sec:Discussions}
        \noindent \textbf{Essence of Flow-based Masks.} We address the essence of flow-based temporal masks in our Video Watermarking. The flow-based masks by selecting key frames method is a powerful tool for video understanding and manipulation, which allows for precise control over specific elements in a video sequence, making it easier to edit and manipulate the video in a variety of ways. We use the clip image-image score of adjacent frames and clip text-image score between the answer and current frame to assess the importance and non-fungibility of our selected frames in Video Watermarking, where a smaller clip image-image score suggests less similarity between a frame and its adjacent frames, and a bigger clip text-image score suggests more similarity between the current frame and the answer of the user input.
        As depicted in Fig.~\ref{fig:flow}, a larger optical flow corresponds to a higher inconsistency between the current frame and its neighboring frames, while containing more information about the answer. This observation suggests that the frames selected using our Video Watermarking are crucial frames in the video, and watermarking them will be more effective.
         \begin{table*}[t]
            \centering
            \caption{Black-box watermarking against VideoChat~\cite{videochat} on the ActivityNet-200~\cite{activitynet} dataset: Comparison of image caption metrics and GPT score for different watermarking types. The sparsity of the temporal mask is set to 0. ${\Delta}$: the mean of the modified pixels. We apply the watermarking video on Video-ChatGPT~\cite{videochatgpt} and directly transfer it to VideoChat~\cite{videochat}.}
            
            \begin{tabular}{c|c|c|c|c|c|c|c|c}
                \hline 
                \multirow{2}{*}{Type} & \multirow{2}{*}{${\Delta}$} & \multicolumn{3}{|c}{Image Caption $\downarrow$} & \multicolumn{2}{|c|}{GPT-3.5 $\downarrow$} & \multicolumn{2}{|c}{GPT-4 $\downarrow$} \\
                \cline{3-9} 
                & & BLEU & ROUGE & CIDEr & Accurate & Score & Accurate & Score \\
                \hline
                Clean  & 0   & 0.0765 & 0.3358 & 0.3379 & 0.35 & 2.87 & 0.28 & 1.79\\
                \hline
                Transfer-based watermarking  & 2    & \textbf{0.0638} & \textbf{0.2492} & \textbf{0.2870} & \textbf{0.08} & \textbf{1.56} & \textbf{0.10} & \textbf{1.32}\\
                \hline
            \end{tabular}
            \label{tab:Black}
        \end{table*}
        
        \noindent \textbf{Transfer-based Black-box Watermarking}
        \label{sec:Black}
        \begin{figure}[ht]
            \centering            
            \includegraphics[width=\linewidth]{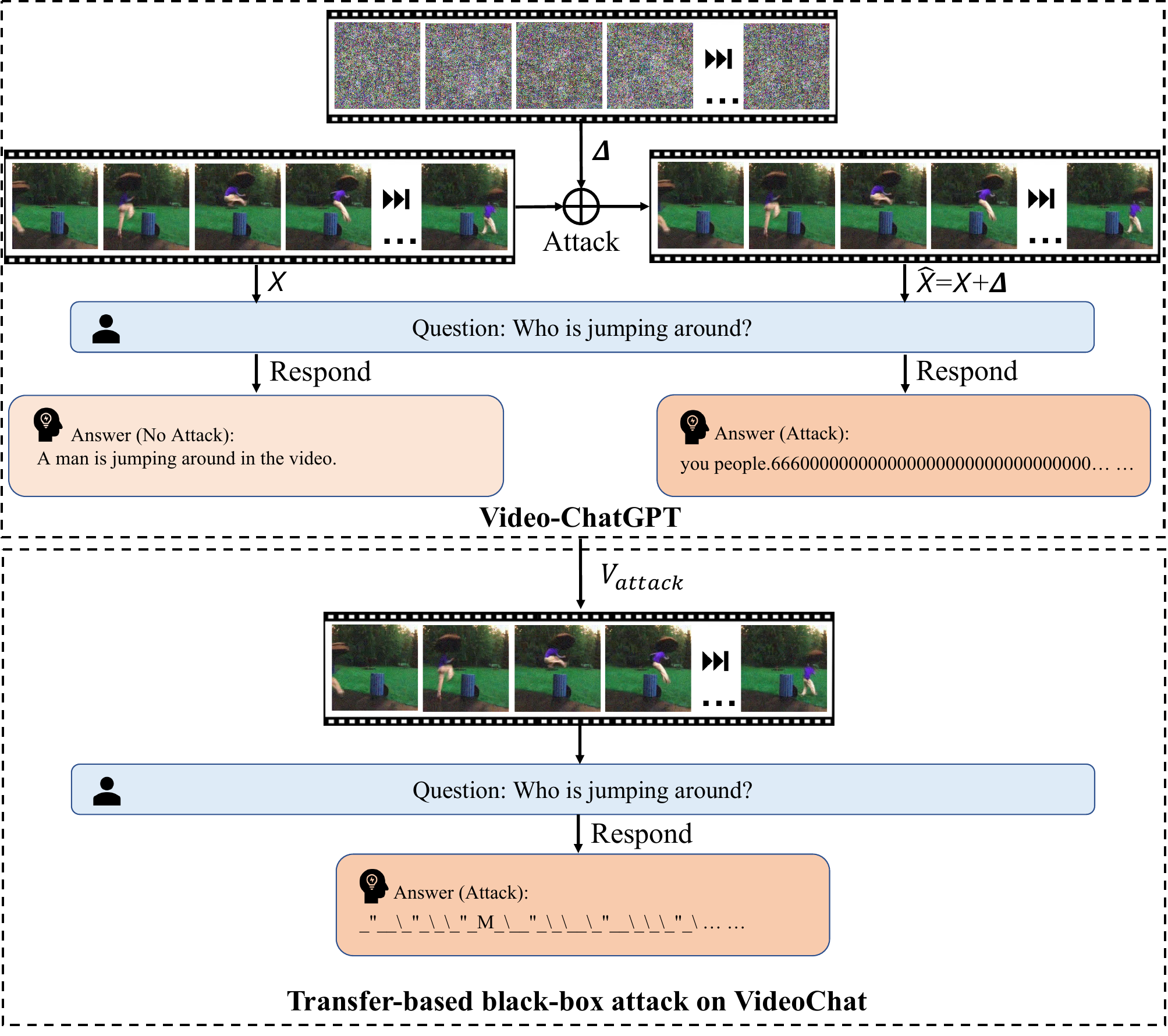}
            \vspace{-1em}
            \caption{Transfer-based black-box watermarking on VideoChat.}
            \label{fig:blackbox}
        \end{figure}

        In addition to white-box watermarking, we have also investigated the transferability of these watermarking. We conduct a black-box watermarking on VideoChat~\cite{videochat}. Specifically, we employ the Video Watermarking method to perform a white-box watermarking on Video-ChatGPT~\cite{videochatgpt}, resulting in the watermarking video $V_{attack}$. This video $V_{attack}$ is then directly used as input for VideoChat~\cite{videochat}, with the experimental results displayed in Table~\ref{tab:Black}. It is evident that the model's answer accuracy decreases significantly. Even without obtaining the gradient of VideoChat~\cite{videochat}, the watermarking is successful, and there are instances of garbled text. The visualization results is shown in Fig.~\ref{fig:blackbox}.

    \subsection{Ablation Studies}
        \noindent \textbf{Loss of Different Modalities.} 
        In Table~\ref{tab:Type}, the video + LLM watermarking outperforms the individual video and LLM watermarking across all metrics, demonstrating the superior performance of the combined approach. This can be attributed to the complementary nature of video and LLM features, which, when targeted simultaneously, leads to a more potent Video Watermarking that effectively disrupts the model's output, resulting in lower scores.

    \subsection{Limitation}
        Our Video Watermarking is primarily concentrated on the digital world, operating under the assumption that input videos are fed directly into the models. However, as technology advances, we anticipate that video-based LLMs will be increasingly deployed in more complex, real-world scenarios. These scenarios could include autonomous driving, where input videos are not pre-recorded but rather captured in real-time from physical environments via cameras. Future research should explore the execution and impact of watermarking in the physical world. This would provide a more comprehensive evaluation of the security of video-based LLMs, contributing to the development of more robust and reliable systems for real-world deployment.
    
\section{Conclusion}
\label{sec:conclusion}
    In this paper, we present Flow-based Video Watermarking, pioneering a new frontier in safeguarding video content against unauthorized exploitation by video-based LLMs. Our experiments demonstrate that with minimal watermarking on less than 20\% of video frames, we can significantly protect video data from misinterpretation and misuse, highlighting the efficacy of our method.
    Our work also sheds light on the broader implications for multi-modal model security. The introduction of our watermarking technique is a testament to the proactive steps necessary to ensure the ethical and secure application of AI technologies in handling sensitive video content.

\section*{Impact Statement} 
    The social impact of our work is significant as it aims to increase public awareness about the concerns associated with the availability of video-based LLMs. The misuse and misinterpretation of video data can lead to serious privacy leakage and misinformation. By highlighting the responsible use and handling of such video information, this work could potentially lead to the development of more secure systems and policies, thereby ensuring the privacy and safety of individuals and communities.

\bibliography{main}
\bibliographystyle{icml2024}

\newpage
\appendix
\onecolumn
\section{Algorithm details}
    Our overall watermarking procedure is described in Algorithm~\ref{alg:1}.
    \begin{algorithm}
        \caption{Video Watermarking on Video-based LLMs Using PGD Optimization}
        \label{alg:1}
        \begin{algorithmic}[1]
        \REQUIRE Clean video $\mathbf{X}$, user input text $Q_{text}$, sparsity $M_{spa}$, video feature extractor $f_\phi(\cdot)$, LLM $g_\psi(\cdot)$, step size $\alpha$, iterations $T$
        \ENSURE Adversarial video $\hat{\mathbf{X}}$
        \STATE Compute video optical flow and obtain flow-based temporal mask $\mathbf{M}_f$ with $M_{spa}$
        \STATE Initialize perturbation $\Delta \gets 0$
        \WHILE{$t < T$}
            \STATE Calculate video features loss $\ell_{video}(Q_{video}, \hat{Q}_{video})$ using Eq.~\ref{loss:video}
            \STATE Calculate LLM features loss $\ell_{LLM}(A_{hidden}, \hat{A}_{hidden})$ using Eq.~\ref{loss:LLM}
            \STATE Update perturbation $\Delta$ using Eq.~\ref{untarget} with step size $\alpha$
        \ENDWHILE
        \STATE Compute adversarial video $\hat{\mathbf{X}} \gets \mathbf{X} + \mathbf{M}_f \cdot \Delta$
        \STATE \textbf{Return:} Adversarial video $\hat{\mathbf{X}}$
        \end{algorithmic}
    \end{algorithm}

\section{Implementation Details}
    \label{sec:Detail}
    \noindent \textbf{Models.} We assess open-source and state-of-the-art video-based LLMs such as Video-ChatGPT~\cite{videochatgpt} and VideoChat~\cite{videochat}, ensuring reproducibility of our results. Video-ChatGPT is a multi-modal model that seamlessly combines a video-adapted visual encoder (CLIP~\cite{clip}) with a LLM, which is proficient in comprehending and generating intricate conversations related to videos. 
    VideoChat integrates video foundation models and large language models via a learnable neural interface.
    \begin{figure}[ht]
        \centering
        \includegraphics[width=0.9\textwidth]{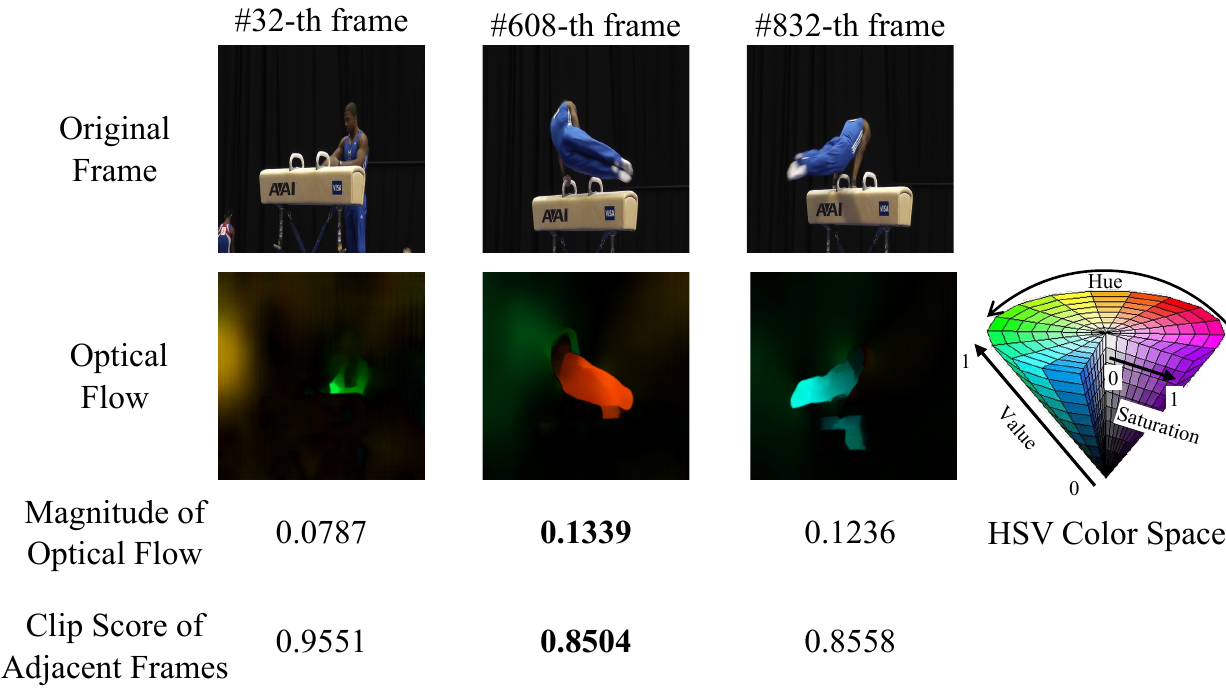}
        \vspace{-1em}
        \caption{Relationship between optical flow and key frames. `Clip Score of Adjacent Frames' describes the similarity between the current frame and its adjacent frames, the smaller this score is the more different the current frame is. \textbf{The frames selected by flow-based masks in our Video Watermarking are key frames in the video.}}
        \label{fig:flow_inter_ori}
    \end{figure}
    
    \noindent \textbf{Flow-based Temporal Mask.} In addition to the statistical analyses presented in the main manuscript, we visualize flow-based methods to enhance interpretability. As depicted in Fig.~\ref{fig:flow_inter_ori}, the brightness represents the magnitude of the optical flow, and the color indicates the motion direction. The motion flow magnitude varies across different frames, with a larger optical flow signifying more significant motion. Our FMM method tends to select the video frames with the largest optical flow as the key frame of the video. In other words, we tend to prioritize frames with substantial motion changes. Furthermore, the frames we select exhibit low similarity with their neighboring frames, indicating their importance.
    
    Algorithm~\ref{alg:top_k_frames} delineates the process of generating the selected set $U$ from the total set $S$. The optical flow is computed using a pre-trained liteflownet~\cite{liteflownet}.
    For the random temporal mask, we generate the selected set $U$ by randomly selecting $K$ elements from the total set $S$. Here, $S$ represents the set of frame indices, $S = \{1, 2, \ldots, T\}$, where $T$ denotes the total number of frames in the video.
    For the sequence temporal mask, we construct the selected set $U$ by sequentially selecting $K$ elements from the total set $S$. In this scenario, $U$ comprises a sequence of frames from total frames $S$, such as $\{1, 2, \ldots, K\}$ or $\{T-K+1, T-K+2, \ldots, T\}$, depending on the selected starting point within $S$.
        
    \begin{algorithm}
        \caption{Select Top K Frames with Maximum Flow}
        \label{alg:top_k_frames}
        \begin{algorithmic}[1]
        \REQUIRE video\_frames $\in T \times C \times H \times W$, $S = \{1, 2, \ldots, T\}$
        \ENSURE U, a subset with $K$ elements within $S$
        \FOR{each frame in video\_frames}
            \STATE Compute optical flow between adjacent frames
        \ENDFOR
        \FOR{each optical flow}
            \STATE Convert optical flow to color and magnitude components
            \STATE Normalize the magnitude component
        \ENDFOR
        \STATE Sort the frames based on the average value of the magnitude component
        \STATE Select the top K frame indices with the highest flow values as the set U
        \end{algorithmic}
    \end{algorithm}

    \noindent \textbf{Datasets.} In line with the Video-ChatGPT~\cite{videochatgpt} methodology, we curate a test set based on the ActivityNet-200~\cite{activitynet} and MSVD-QA~\cite{MSVDQA} datasets, featuring videos with rich, detailed descriptive captions and associated question-answer pairs obtained from human annotations. Utilizing this test set to generate adversarial examples, we effectively and quantitatively assess the adversarial robustness of video-based LLMs.

    \noindent \textbf{Experimental setups.} For evaluation, we design three spatial baselines, including videos with random perturbations, black videos with all pixel values set to 0, and white videos with all pixel values set to 1. In addition, we compare our proposed flow-based temporal mask with two straightforward temporal mask methods, serving as temporal mask baselines: the sequence temporal mask and the random temporal mask. Specifically, the sequence temporal mask consists of a continuous sequence of frame indices, while the random temporal mask comprises a randomly chosen sequence of frame indices. We utilize a variety of evaluation metrics to assess the robustness of the models.
    CLIP~\cite{clip} score characterizes the semantic similarity between the adversarial answer and the ground-truth answer. A lower CLIP score signifies a lower semantic correlation between the adversarial answer and the ground-truth answer, indicating a more effective attack. Various Image Captioning metrics such as BLEU~\cite{bleu}, ROUGE-L~\cite{rouge}, and CIDEr~\cite{cider} are used to evaluate the quality of the adversarial answer generated by the model. A lower score corresponds to a more effective attack.
    BLEU measures the overlap of n-grams between the generated and reference captions. ROUGE-L computes the longest common subsequence between them, reflecting their sentence-level similarity. CIDEr emphasizes the importance of semantically meaningful words in the captions.
    Following Video-ChatGPT~\cite{videochatgpt} and Video-LLaMA~\cite{videollama}. We also employ an evaluation pipeline using the GPT-3.5 and GPT-4 models. The pipeline employs GPT to assign a score from 1 to 5, evaluating the similarity between the output sentence and the ground truth, and a binary score (0 or 1) to measure its accuracy.

\section{Additional Experimental Results}
\begin{figure}[t]
            \centering
            \includegraphics[width=0.7\linewidth]{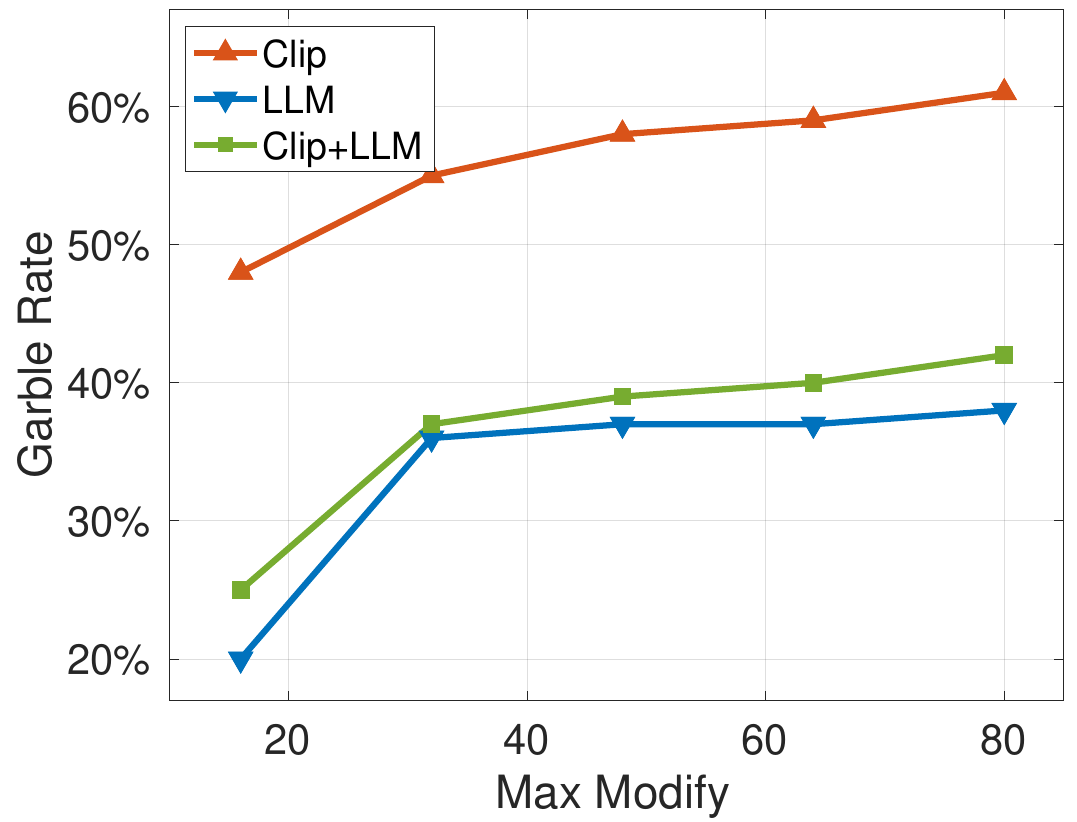}
            \vspace{-1em}
            \caption{Comparison of different types of watermarking on the garbling rate. Max Modify denotes the maximum pixel value that can be modified, while the Garble Rate represents the percentage of responses that are garbled.}
            \label{fig:garble2}
        \end{figure}

    \noindent \textbf{Garbling Effect.}
        Intriguingly, our proposed Video Watermarking induces garbling in the model output, while the other three watermarking methods do not cause such distortion. This suggests that the Video Watermarking not only diminishes the model's cue information but also prompts the model to hallucinate. Furthermore, as shown in Fig.~\ref{fig:garble2}, we analyse the number of successfully attacked Video-ChatGPT in ActivityNet-200~\cite{activitynet}, and find that video loss is more effective in inducing garbled contents.

    \begin{figure}[t]
        \centering
        \includegraphics[width=\linewidth]{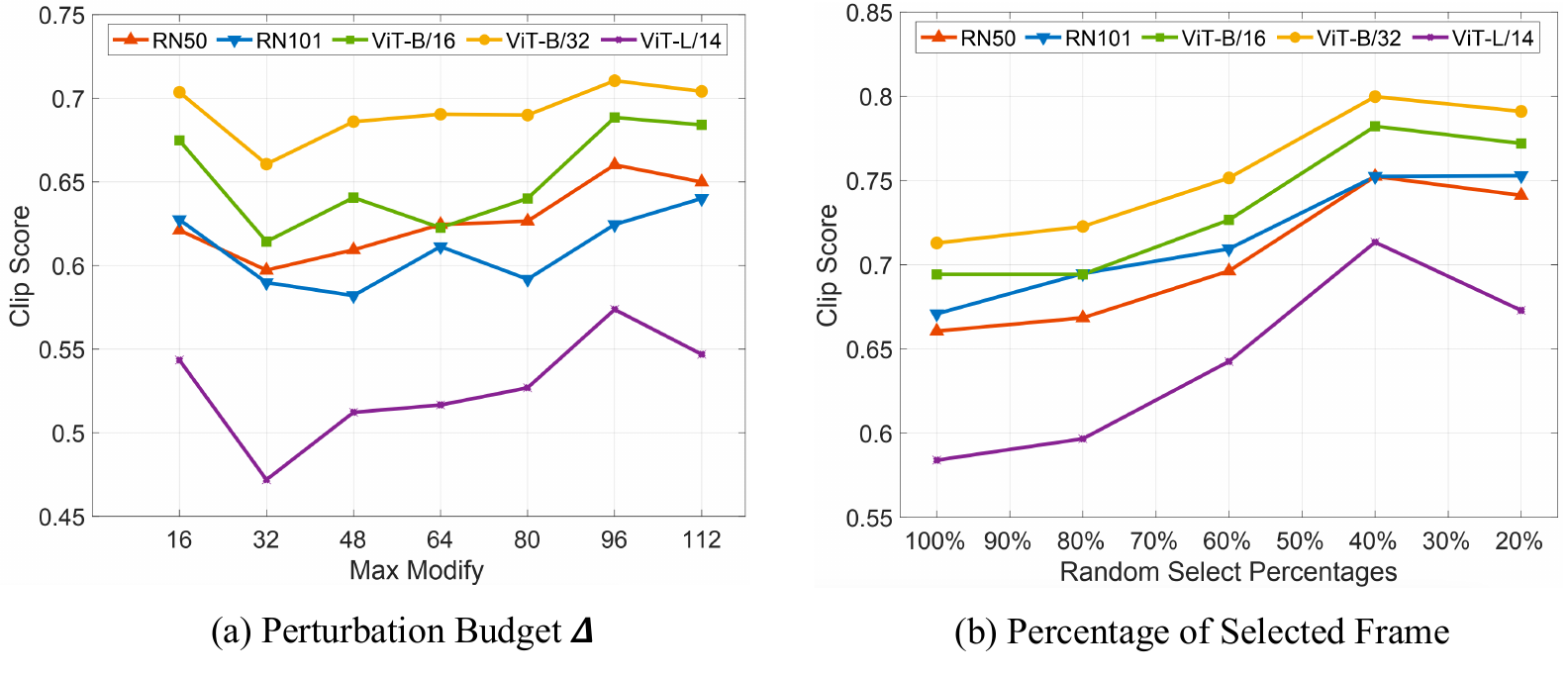}
        \vspace{-1em}
        \caption{Ablation Studies of different watermarking settings. (a) Comparison of different max modify pixels. (b) Comparison of different random select percentages.}
        \label{fig:merge}
    \end{figure}
    \noindent \textbf{Perturbation Budget $\Delta$.} We compared the effects of different $\Delta$. It's important to note that the mean of $\Delta$, constrained by the sparsity loss, remains consistent across various levels of maximum modification $\Delta_{max}$. As in Fig.~\ref{fig:merge} (a), $\Delta = 32$ performs best. On the one hand, the larger $\Delta$ is, the greater the potential modification of individual pixels, but on the other hand, due to the sparse loss, the amount of pixels that can be modified with a larger $\Delta$ becomes smaller. Therefore, $\Delta = 32$ seems to be a better trade-off. 
    \noindent \textbf{Percentage of Selected Frames.} In some instances, while we launch watermarking on all video frames, video-based LLMs only randomly sample a subset of the video frames. As demonstrated in Fig.~\ref{fig:merge} (b), the potency of the watermarking escalates with an increasing number of sampled video frames. Interestingly, even when a minor fraction (40\%, 20\%) of video frames are sampled, we observe a substantial decline in the Clip score relative to the baseline. Surprisingly, a 20\% sampling rate seems to yield superior results than a 40\% rate. We speculate this unexpected outcome could be due to inherent fluctuations when a limited number of video frames are sampled, coupled with our stream-based approach that assures a certain minimum watermarking effectiveness. To make this engagement more compelling, we intend to delve deeper into this phenomenon with additional experiments in future studies.

    \begin{table}[t]
        \small
        \centering
        \caption{Video watermarking against Video-ChatGPT on the ActivityNet-200 dataset. seq: sequence temporal mask. random: random temporal mask. Ours: our flow-based temporal mask. ${M}_{spa}$: Sparsity of temporal mask.}
        \begin{tabular}{c|c|c|c|c|c|c|c|c|c|c|c|c}
        \hline
        \multirow{2}{*}{ Metrics } & \multicolumn{3}{|c|}{ ${M}_{spa} = 20\%$ } & \multicolumn{3}{|c|}{ ${M}_{spa} = 40\%$ } & \multicolumn{3}{|c|}{ 
        ${M}_{spa} = 60\%$} & \multicolumn{3}{|c}{ ${M}_{spa} = 80\%$ } \\
        \cline{2-13} & seq & random & Ours & seq & random & Ours & seq & random & Ours & seq & random & Ours \\
        \hline
        RN50 & 0.7500 & 0.7142 & 0.6821 & 0.7578 & 0.7559 & 0.7460 & 0.7583 & 0.7671 & 0.7510 & 0.7715 & 0.7813 & 0.7349 \\
        RN101 & 0.7568 & 0.7251 & 0.7085 & 0.7500 & 0.7588 & 0.7559 & 0.7764 & 0.7769 & 0.7500 & 0.7788 & 0.7827 & 0.7637 \\
        \hline
        BLEU & 0.1572 & 0.1590 & 0.1527 & 0.1606 & 0.1751 & 0.1485 & 0.1900 & 0.1894 & 0.1830 & 0.2000 & 0.1957 & 0.1940 \\
        ROUGE-L & 0.4073 & 0.3996 & 0.4109 & 0.4323 & 0.4458 & 0.4053 & 0.4767 & 0.4727 & 0.4713 & 0.4779 & 0.4658 & 0.4731 \\
        \hline
        GPT3.5 & 2.02 & 2.01 & 2.00 & 2.45 & 2.24 & 2.22 & 2.64 & 2.63 & 2.62 & 2.76 & 2.78 & 2.75 \\
        GPT4 & 1.09 & 1.08 & 1.07 & 1.35 & 1.34 & 1.33 & 1.63 & 1.63 & 1.61 & 1.99 & 1.58 & 1.54 \\
        \hline
        \end{tabular}
        \label{tab:TemporalMask}
    \end{table}
    Table~\ref{tab:TemporalMask} compares different mask ratios (sparsity) and temporal mask approaches. Our Video Watermarking outperforms the other two approaches across various sparsity levels, indicating the effectiveness of our proposed flow-based temporal mask, which leverages the concept of maximum flow prioritization. It demonstrates its potency in the realm of video-based LLM and its ability to maximize the extraction of video information.
    In the bottom of Table~\ref{tab:TemporalMask}, Video Watermarking consistently outperforms the other two approaches across various sparsity levels, resulting in a more significant reduction of GPT scores and accuracies.
    \begin{table}
        \centering
        \caption{Comparison of different watermarking types on Clip Score and Image Captioning Metrics. video: represents watermarking targeting video features, LLM: represents watermarking targeting LLM features, and video + LLM: represents combined watermarking on both video and LLM features. Lower scores indicate better watermarking performance.}
        \begin{tabular}{c|c|c|c|c|c|c|c|c}
        \hline
        \multirow{2}{*}{ Type } & \multicolumn{5}{|c|}{ Clip Score $\downarrow$} & \multicolumn{3}{c}{ Image Captioning $\downarrow$} \\
        \cline { 2 - 9 } & RN50 & RN101 & ViT-B/16 & ViT-B/32 & ViT-L/14 & BLEU & ROUGE-L & CIDEr \\
        \hline Clean & 0.7817 & 0.7827 & 0.8096 & 0.8115 & 0.7231 & 0.2029 & 0.4820 & 1.6364 \\
        \hline video & 0.7403 & 0.7524 & 0.7690 & 0.7744 & 0.6811 & 0.1975 & 0.4345 & 1.5862 \\
        \hline LLM & 0.7153 & 0.6904 & 0.7334 & 0.7515 & 0.6387 & 0.1042 & 0.3226 & 0.8350 \\
        \hline video + LLM & \textbf{0.6491} & \textbf{0.6060} & \textbf{0.6836} & \textbf{0.6968} & \textbf{0.5566} & \textbf{0.0230} & \textbf{0.1585} & \textbf{0.1601} \\
        \hline
        \end{tabular}
        \label{tab:Type}
    \end{table}

\end{document}